\documentclass[preprint,3p,times,twocolumn,numbers]{elsarticle}

\usepackage{amssymb}
\usepackage{amsmath}
\usepackage{amsthm}
\usepackage{caption}
\usepackage{subcaption}
\usepackage{setspace}
\usepackage{physics}
\usepackage{multirow}
\usepackage[table]{xcolor}
\usepackage{imakeidx}
\usepackage{url}

\usepackage{lineno}
\newcommand{\dLITE}[0]{$\partial$LITE}

\definecolor{yesgreen}{RGB}{200,255,200}
\definecolor{nored}{RGB}{255,200,200}
\definecolor{maybe}{RGB}{255,255,180}

\journal{Acta Astronautica}

\begin{document}

\begin{frontmatter}



\title{$\partial$LITE: Differentiable Lighting-Informed Trajectory Evaluation for On-Orbit Inspection} 

\author[label1]{Jack Naylor}
\author[label1,label2]{Raghav Mishra}
\author[label1]{Nicholas H.\,Barbara}
\author[label1,label2]{Donald~G.\,Dansereau}

\affiliation[label1]{organization={Australian Centre for Robotics and School of Aerospace, Mechanical and Mechatronic Engineering, The University of Sydney},
            postcode={2006},
            state={NSW},
            country={Australia}}

\affiliation[label2]{organization={Australian Robotic Inspection and Asset Management Hub (ARIAM), The University of Sydney},
            postcode={2006},
            state={NSW},
            country={Australia}}



\begin{abstract}
Visual inspection of space-borne assets is of increasing interest to spacecraft operators looking to plan maintenance, characterise damage, and extend the life of high-value satellites in orbit. The environment of Low Earth Orbit (LEO) presents unique challenges when planning inspection operations that maximise visibility, information, and data quality. Specular reflection of sunlight from spacecraft bodies, self-shadowing, and dynamic lighting in LEO significantly impact the quality of data captured throughout an orbit. This is exacerbated by the relative motion between spacecraft, which introduces variable imaging distances and attitudes during inspection. Planning inspection trajectories with the aide of simulation is a common approach. However, the ability to design and optimise an inspection trajectory specifically to improve the resulting image quality in proximity operations remains largely unexplored. In this work, we present \dLITE{}, an end-to-end differentiable simulation pipeline for on-orbit inspection operations. We leverage state-of-the-art differentiable rendering tools and a custom orbit propagator to enable end-to-end optimisation of orbital parameters based on visual sensor data. \dLITE{} enables us to automatically design non-obvious trajectories, vastly improving the quality and usefulness of attained data. To our knowledge, our differentiable inspection-planning pipeline is the first of its kind and provides new insights into modern computational approaches to spacecraft mission planning.

\end{abstract}





\end{frontmatter}



\section{Introduction}

As the number and value of assets in space rapidly increases, there is a growing need for reliable and efficient visual inspection capabilities in orbit. Spacecraft operators can use inspection data to assess damage, plan maintenance, and extend the operational lifespan of high value satellites. However, conducting visual inspections in Low Earth Orbit (LEO) presents a unique set of challenges. The dynamic lighting conditions caused by specular reflections, self-shadowing, and the rapid orbital motion of spacecraft can significantly degrade image quality and limit the visibility of key features~\citep{amaya2024visual}. Moreover, the relative motion between inspection and target spacecraft introduces variability in imaging distance and orientation, further complicating data acquisition.

\begin{figure}[htbp]
    \centering
    \includegraphics[width=\linewidth]{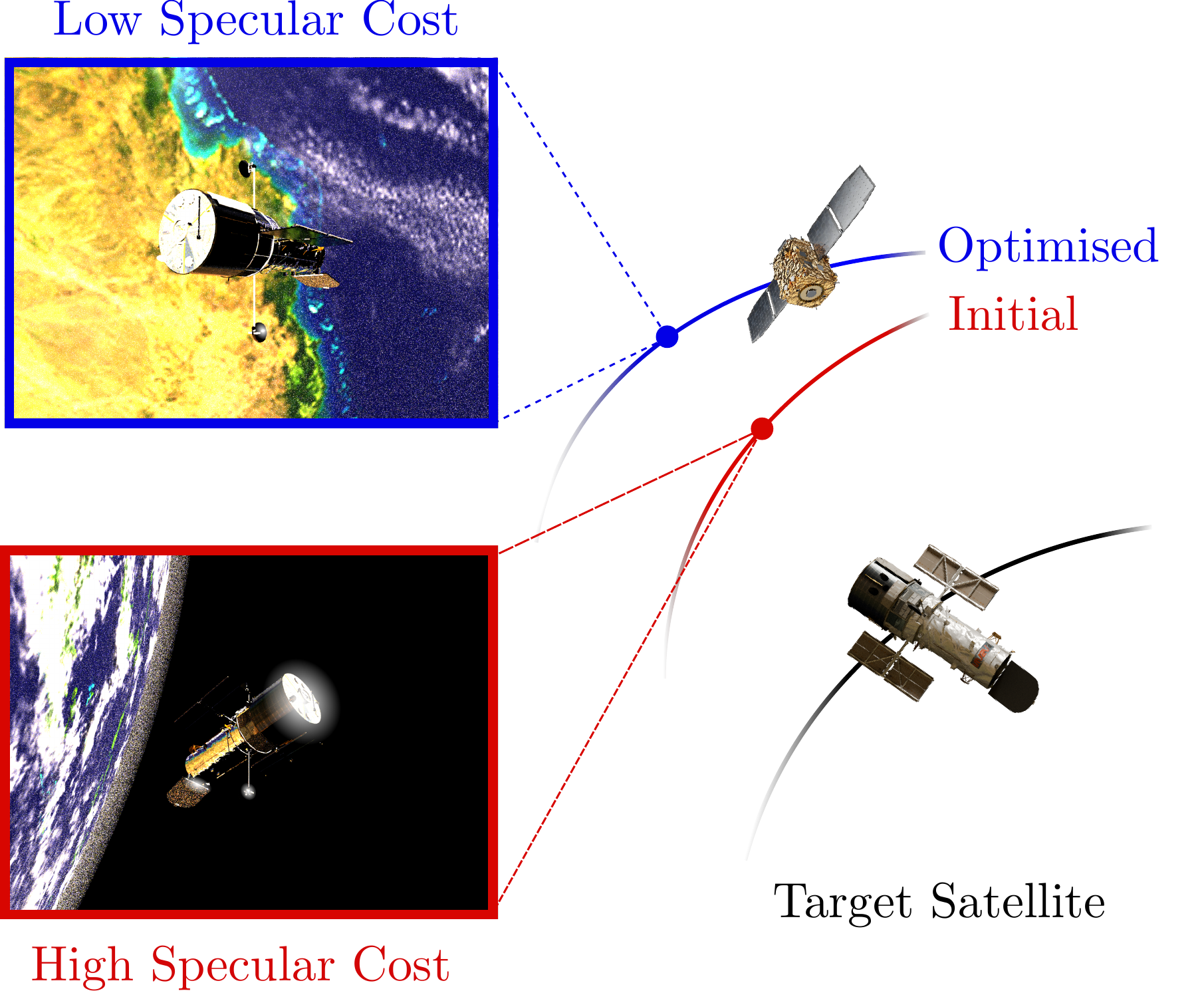}
    \caption{\dLITE{} is an end-to-end differentiable simulator for on-orbit inspection which produces non-trivial trajectories based on visual costs. We optimise orbital parameters for passive inspection trajectories that yield high-quality images of a target satellite by minimising specular reflections seen by the sensor.}
    \label{fig:fig1}
\end{figure}

Historically, on-orbit inspection and servicing of spacecraft in LEO has been performed manually to return high-value assets to service~\citep{mcmahan1984repairing, NASA1994_STS61_MissionReport}, and to extend mission life and capability~\citep{, NASA1997_STS82_MissionReport, NASA1999_STS103_MissionReport, NASA2002_STS109_MissionReport, NASA2009_STS125_MissionReport}. Autonomous on-orbit inspection, however, is a more recent capability. Demonstrator missions for proximity operations like Orbital Express~\citep{friend2008orbital}, precision formation-flying missions like PROBA-3~\citep{tiraplegui2019proba3}, and emerging servicing missions like Northrup Grumman's Mission Extension Vehicle (MEV)~\citep{northropgrumman_mev_overview}, have demonstrated key capabilities relied upon to achieve on-orbit servicing. Planning informative inspections which attain high-fidelity visual data to inform these servicing operations remains a challenge.

Inspection and rendezvous missions in high Earth orbits such as Geostationary Orbit (GEO)~\citep{pyrak2021mev_rpo_imagers} benefit from sustained illumination and often slow relative dynamics, allowing for predictable imaging of target spacecraft. In contrast, inspection in LEO has varying imaging geometries owing to stronger orbital perturbations, highly dynamic lighting environments, and rapid relative motion which vastly complicate the inspection of satellites. Techniques for long-range imaging of satellites in LEO~\citep{allworth2024use} are more resilient to such complicating factors, but they are unable to characterise damage and component level failure in the resolution needed to plan servicing operations. To enable related operations like docking and state estimation of satellites, alternate imaging modalities like event cameras~\citep{gentil2025mixing, jawaid2023towards} have been proposed to handle imaging in changing lighting conditions with High Dynamic Range (HDR). However, such methods require substantial processing to provide the visual information to perform change detection, damage characterisation, and 3D reconstruction for repair and servicing operations compared to traditional imaging.

While simulation-based trajectory planning is common practice for on-orbit inspection, current methods overlook the impact of lighting conditions on the quality of imaging data. To address this gap, we introduce \dLITE{}---a novel, end-to-end differentiable simulation pipeline specifically designed for visual on-orbit inspection planning. By integrating state-of-the-art differentiable rendering techniques with a differentiable orbit propagator, \dLITE{} enables the optimisation of inspection trajectories based on visual measurements, allowing the user to refine orbital paths to improve relative lighting conditions, and ultimately the fidelity of collected data as shown in Figure~\ref{fig:fig1}. To our knowledge, \dLITE{} represents the first fully differentiable framework for planning visual inspection trajectories in space. 

Our main contributions are as follows:
\begin{itemize}
    \item We propose a fully differentiable simulator, combining a custom, fully-differentiable implementation of the SGP4 orbital propagator~\citep{vallado2006revisiting} with a photometrically accurate differentiable renderer built on the Mitsuba 3 rendering engine~\citep{jakob2022mitsuba3};
    \item We demonstrate end-to-end optimisation of inspection orbits to minimise visual costs associated with blinding specular reflections using gradient-based optimisation; and
    \item We show that our optimised orbits improve other proxy metrics for close-range satellite operations such as the number of visual features detected over the trajectory.
\end{itemize}

We envisage that \dLITE{} will enable tailored trajectory planning of favourable trajectories for on-orbit inspection missions. More broadly, we anticipate that our pipeline will underpin the development of future planning approaches that are \textit{sensitive} and consider complex visual appearances that include specularity~\citep{mishra2024apaware}, shadowing~\citep{kitamura2024shadow}, and self-occlusion. Such approaches consider the quality of visual information captured, which is critical under the time, resource, and cost constraints of satellite missions compared to traditional visibility~\citep{zou2018optimal, xue2024neural} and coverage-based planning methods~\citep{strimel2014coverage, galceran2013survey}.

\section{Background}
Existing pipelines for simulation-based inspection planning typically focus only on simulating sensor data~\citep{felicatti2023hysim}, dataset generation for pose estimation and navigation~\citep{montalvo2024spin}, or existing mission trajectories without optimisation~\citep{pajusalu2022sispo, li2024modular}. While tools such as SISPO~\citep{pajusalu2022sispo} do include complex visual effects such as optical distortion and rendering of custom point spread functions, they do not allow direct computational design of orbits. \dLITE{} combines orbital simulation with a high-fidelity visual simulation of satellite observations. To enable direct optimisation of orbital parameters, we develop \dLITE{} with fully differentiable photorealistic rendering and orbital propagation. We compare the capabilities of prior works with those of \dLITE{} in Table~\ref{tab:literature}.

\begin{table}[ht]
    \centering
        \caption{Comparison of visual simulation frameworks for on-orbit inspection missions. \dLITE{} is an end-to-end differentiable, photometrically accurate simulator which can produce photorealistic renderings of observations from an inspection satellite. Green indicates full capabilities, yellow indicates partial capability, and red indicates no capability.}\label{tab:literature}
    \begin{tabular}{l
                p{0.35cm}
                p{0.35cm}
                p{0.35cm}
                p{0.35cm}
                p{0.35cm}}
        Name &
\rotatebox{60}{\scriptsize Diff. Orbits} &
\rotatebox{60}{\scriptsize Diff. Rendering} &
\rotatebox{60}{\scriptsize  Photometric} &
\rotatebox{60}{\scriptsize Optical Effects} &
\rotatebox{60}{\scriptsize Photorealistic} \\
\hline
        ALL-STAR~\citep{li2024modular} & \cellcolor{nored} & \cellcolor{nored} & \cellcolor{nored} & \cellcolor{nored} & \cellcolor{maybe}\\
        SPIN~\citep{montalvo2024spin} & \cellcolor{nored} & \cellcolor{nored} &\cellcolor{nored} & \cellcolor{nored} & \cellcolor{yesgreen} \\
        SISPO~\citep{pajusalu2022sispo} & \cellcolor{nored} & \cellcolor{nored} &\cellcolor{nored} & \cellcolor{yesgreen} & \cellcolor{yesgreen}\\
        HySim~\citep{felicatti2023hysim} & \cellcolor{nored} & \cellcolor{maybe} & \cellcolor{yesgreen} & \cellcolor{maybe} & \cellcolor{yesgreen}\\
        \dLITE{} (Ours)  &  \cellcolor{yesgreen} & \cellcolor{yesgreen} &  \cellcolor{yesgreen} & \cellcolor{maybe} & \cellcolor{yesgreen}\\
        \hline
    \end{tabular}

\end{table}

In the following subsections, we provide a brief overview of relevant background on Non-Earth Imaging (NEI), orbital trajectory planning, differentiable programming, and differentiable scene rendering, which form the core elements of \dLITE{}.

\subsection{Non-Earth Imaging}
Recent advances in NEI have allowed for high quality classification and characterisation of space objects using visual imagery. Private organisations such as Vantor~\citep{vantor2025NEI} and HEO ~\citep{heo_nei_whitepaper_2025} have demonstrated these capabilities commercially using fly-by NEI, whereby satellites are tasked with capturing high resolution images of target objects. This data allows for visual identification of objects from a distance, including satellites, discarded rocket bodies, and other unknown objects in LEO~\citep{allworth2024use}.

\begin{figure}[ht]
    \centering
    \begin{subfigure}[b]{0.48\linewidth}
        \centering
        \includegraphics[width=\linewidth]{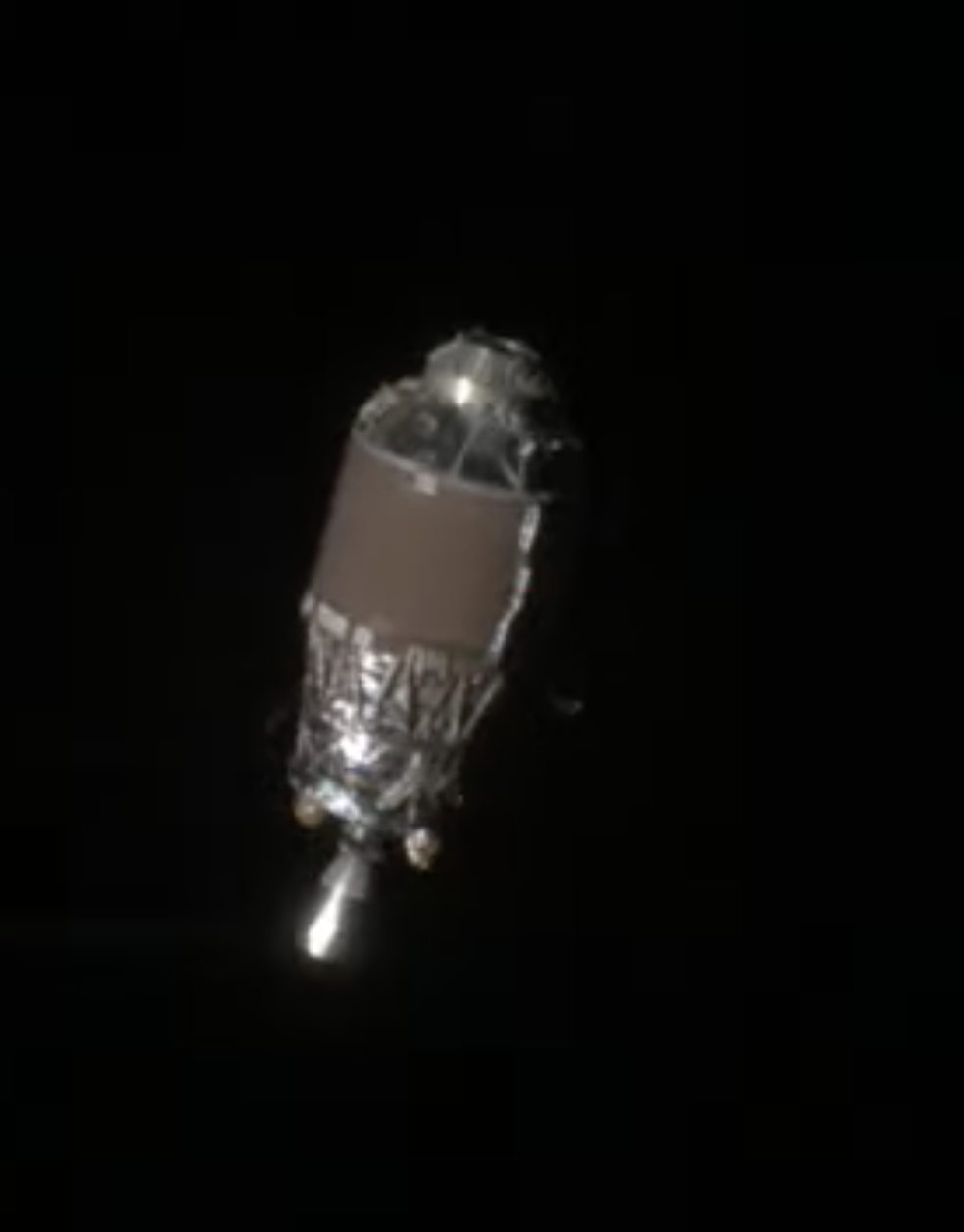}
    \end{subfigure}
    \begin{subfigure}[b]{0.48\linewidth}
        \centering
        \includegraphics[width=\linewidth]{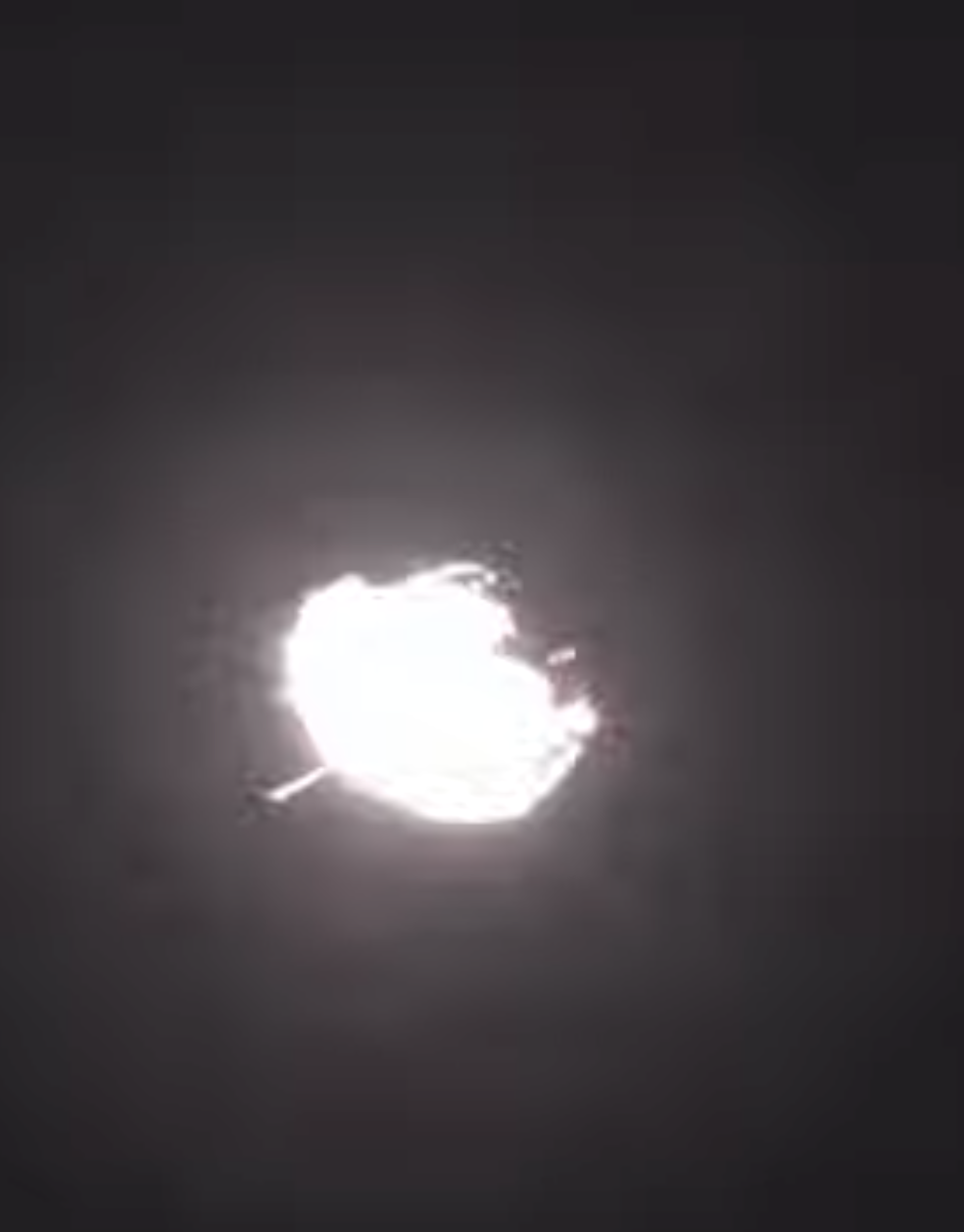}
    \end{subfigure}
    \caption{Images of an H-IIA upper stage rocket body captured by Astroscale's ADRAS-J mission~\citep{JAXA2024ADRASJ}. The mission was able to capture high-quality images of the rocket body over much of its inspection orbit (left). However, high dynamic range content, saturation and sensor bloom due to specular reflections make some images uninformative (right).}
    \label{fig:astroscale}
\end{figure}

Most recently, Astroscale's ADRAS-J mission~\citep{JAXA2024ADRASJ} has demonstrated sustained, high-resolution inspection of non-cooperative debris in close proximity. By taking images of an upper stage rocket body of the H-IIA launch system, they provided insight into features of the target craft such as its attitude dynamics, and ultraviolet degradation of its thermal insulation, thereby informing future missions to capture and de-orbit the debris. However, owing to the dynamic illumination and relative motion in orbit, the quality of captured imagery varies significantly during an inspection orbit. At certain points in the orbit, blinding reflections and HDR conditions appear, which make it  difficult to extract useful insight from the imagery (Figure~\ref{fig:astroscale}). 

In this work, we consider similar close proximity inspection missions, and we focus on optimising orbital trajectories to maximise the quality of \textit{all} images captured during a designated inspection period, thereby avoiding events such as that shown in Figure~\ref{fig:astroscale}.

\subsection{Orbital Trajectory Planning}
Orbital trajectory optimisation is widely used when (for example) planning interplanetary trajectories \citep{rasotto2016differential,addis2011global,vasile2010analysis}, navigating through complex gravitational zones with many interacting perturbations \citep{russell2012survey}, for manoeuvres transferring between orbits \citep{prussing2018optimal}, and for managing swarms of inspection spacecraft
\citep{bernhard2020coordinated}. 
It makes use of a rich literature of trajectory optimisation tools developed for terrestrial applications such as those in robotics \citep{LaValle2006,jackson2021planning,howell2019altro}. In the context of astrodynamics, the trajectory of a spacecraft is numerically optimised by computing the thrust forces and/or attitude for a given objective function to be maximised.

For inspection tasks, one must design orbital trajectories that maintain an appropriate distance to and view of the target satellite. A typical strategy when designing close-proximity inspection orbits for a given target satellite is to make use of the periodic relative motion that naturally occurs between two spacecraft in similar orbits.
For example, if the target is in a circular orbit, and the orbit of the inspection spacecraft is identical except for a small change in eccentricity, then the relative motion of the inspector with respect to the target is elliptical (i.e., a football orbit \citep{woffinden2004Onorbitsatellite}). 
This allows the inspector to view the target from a range of different angles simply by leveraging the passive orbital dynamics of the two bodies.

Common inspection orbits \citep{woffinden2004Onorbitsatellite} are manually chosen to maximise coverage of the target---or a specific site of interest on the target---in the relative frame, ensuring that sufficient image data can be captured over multiple passes. If, however, one wishes to impose more sophisticated design criteria on the inspection orbit, then more sophisticated methods are required.

\subsection{Differentiable Programming}
Computational design of orbits using gradient-based optimisation algorithms can be challenging due to the difficulty of expressing the derivatives of an objective functions with respect to the design variables. This has traditionally hampered the use of standard gradient-based algorithms in astrodynamics, and has resulted in the use of black-box meta-heuristic algorithms such as genetic algorithms \citep{Shirazi_Ceberio_Lozano_2017}. An alternate approach to addressing this difficulty is to leverage modern advances in differentiable programming \citep{blondel2024elements}. Differentiable programming languages~\citep{bezanson2017julia} and libraries~\citep{jax2018github} allow users to write code that is automatically differentiable---that is, gradients of functions and other operations can be automatically and efficiently computed without the user having to explicitly define them. Writing algorithms in a differential programming package enables a user to optimise inputs, outputs, or algorithm parameters for computational design, model learning, uncertainty propagation, sensitivity analysis, and much more. This simple and flexible framework makes it easy for users to optimise arbitrary objective functions with first-order optimisation schemes (e.g., gradient descent), and has led to its widespread use in optimisation~\citep{howell2019altro,howell2022calipso} and machine learning~\citep{sutton2018reinforcement,paszke2019pytorch}. More recently, differentiable physics simulators~\citep{howell2022dojo,freeman2021brax,acciarini2024closing} and renderers~\citep{NimierDavidVicini2019Mitsuba2, kato2020differentiable} have emerged, allowing for gradient propagation through complex physical environments and sensor models.

\subsection{Differentiable Rendering for Downstream Tasks}
Differentiable rendering engines are able to render a 3D scene and calculate gradients and Jacobian-vector products of the output with respect to scene parameters. This allows, for example, one to optimise for parameters in the scene such as camera or object poses, textures, surface normals, and more using gradient based optimisation.  Ray-based differentiable rendering pipelines like Mitsuba~3 \citep{jakob2022mitsuba3} enable complex lighting effects like interreflection, occlusion, self-shadowing, and scattering to be represented. These effects are physically based~\citep{pharr2023physically}, unlike in rasterisation-based methods~\citep{Laine2020diffrast}. HDR environments with reflective scene elements, like those those found in space~\citep{jawaid2025event, gentil2025mixing}, produce complex visual appearances which are most accurately represented with physically-based ray tracing.

Increasingly in robotics, creating data for training agents with methods such as reinforcement learning \citep{sutton2018reinforcement} requires not only physically accurate dynamics~\citep{todorov2012mujoco} but also visually accurate scenes~\citep{drake, mittal2023orbit} which minimise the domain gap between simulation and deployment in downstream tasks. In particular, differentiable photorealistic representations of a robot's environment enabled by new scene representations~\citep{irshad2024neuralfieldsroboticssurvey} have enabled downstream tasks from pose optimisation, odometry and trajectory planning~\citep{michaux2025let, andreu2025foci}. We take a similar approach in this work for on-orbit inspection.

\section{Method}

\begin{figure*}[ht]
    \centering
    \includegraphics[width=\linewidth]{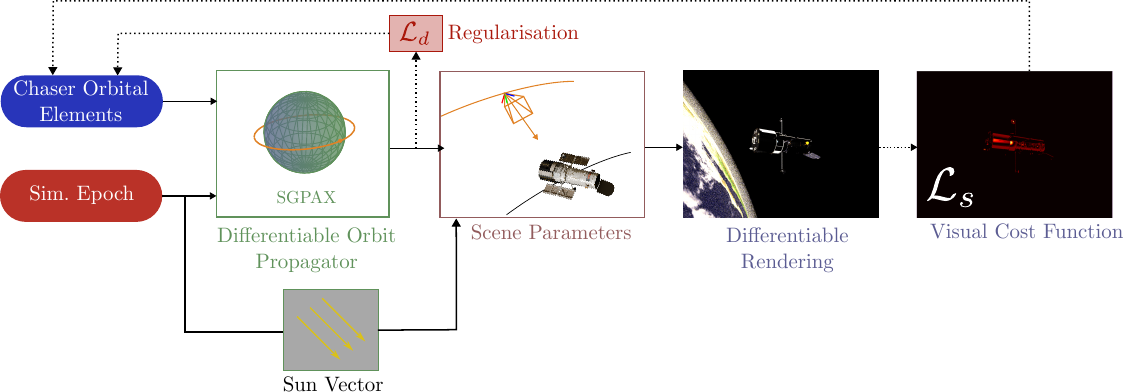}
    \caption{An overview of the \dLITE{} pipeline. Given a TLE for the chaser satellite and a simulation epoch, we compute a time-varying sun vector $\hat{\mathbf{l}}$, and satellite position at time $t$. The scene parameters are updated to produce rendered inspection imagery. From these observations we formulate a visual cost $\mathcal{L}_s$ capturing the specularity in the scene to backpropagate gradients through the pipeline (dotted lines).}
    \label{fig:pipeline}
\end{figure*}

We have created a simulation environment that combines differentiable rendering with a differentiable orbit propagator, allowing for high fidelity ray-traced renderings with photorealistic appearance based on accurate models of orbital dynamics. Combining both of these components, our \dLITE{} pipeline (Figure~\ref{fig:pipeline}) can perform optimisation of orbital elements from costs associated with visual observations to produce ``non-obvious'' relative trajectories that are bespoke to the target satellite and would be difficult to design by hand.

\subsection{Trajectory Optimisation Approach} \label{sec:opt_approach}
Leveraging \dLITE{}, we pose a visual inspection planning problem in a trajectory optimisation framework. We assume the target spacecraft is cooperative, its orbital parameters are static and known, and that it has an accurate 3D model available for use during planning. We directly optimise a set of orbital parameters for a chaser spacecraft as decision variables to minimise a cost on a visual imaging model of the spacecraft. 

As the imaging process is continuous, to make optimisation tractable, we discretise the trajectory by sampling evenly-spaced poses over a time horizon. We consider $N$ evenly spread snapshots over a time period $t\in [0, T]$. 
The trajectory optimisation problem to compute optimal orbital parameters can be written as
\begin{align}
\begin{split}
       \min_{o_c}\quad \sum_{i=0}^N \lambda_S\mathcal{L}&_S(\mathcal{I}(t_i))+\lambda_d\mathcal{L}_d(r_c(t_i), r_t(t)) \\
    \text{s.t. } \mathcal{I}(t_i) & = \texttt{Render}(r_c(t_i), r_t(t_i), t_i) \\
    r_c(t) &= \texttt{SGP4}(t_i, o_\text{c}) \\
    r_t(t) &= \texttt{SGP4}(t_i, o_\text{t}) \\
    t_i &= i\frac{T}{N}
\end{split}
\end{align}
where $o_c$ is the set of chaser orbital elements to optimise over, $\texttt{SGP4}(\cdot)$ is our orbit propagator (Section \ref{sec:difforbit}), $\texttt{Render}(\cdot)$ is our renderer (Section \ref{sec:render}), $r_c$ and $r_t$ are the target positions of the chaser and target satellite respetively, and $\mathcal{L}_S$ and $\mathcal{L}_d$ are our specular intensity cost and a distance regulariser cost, respectively (Section \ref{sec:vis_cost}). $T$ is a period of time we want to optimise over, and $\lambda_S$ and $\lambda_d$ are weights that trade off between the two cost terms.

We perform gradient-based optimisation using the \texttt{Adam} optimiser with a learning rate of $4\times10^{-6}$ to optimise the cost. We optimise for $200$ iterations and empirically find it converges in less than approx. $150$ iterations with our choice of cost functions and optimisation parameters. For our experiments, we choose $o_c=\{i_c, M_c, e_c\}$ which are the chaser's inclination, mean anomaly, and the eccentricity, respectively. However, any standard orbital parameters that are used as input for orbit propagation can be chosen if desired. 

We initialise the inspection craft in an identical orbit to the target satellite except for a small shift in the initial mean anomaly to put it in a periodic circular relative orbit to the target (as each target is already in an approximately circular orbit). We initialise the mean anomaly of the chaser, $M_c$, in terms of the target mean anomaly, $M_t$ as
\begin{gather}
    M_c = M_t - \frac{d}{a_t},
\end{gather}
where $d$ is the radius of the initial relative circular motion, and $a_t$ is the semi-major axis of the target's orbit.
Note that \dLITE{} is compatible with any initial inspection orbit---we simply chose a circular orbit for illustrative purposes. A review of other common inspection orbits can be found in \citep{woffinden2004Onorbitsatellite}.

\subsection{Differentiable Orbit Propagation}\label{sec:difforbit}
Using differentiable programming, along similar lines as \citep{acciarini2024closing}, we implement \texttt{sgpax}---an implementation of the SGP4 orbit propagator \citep{spacetrack3}, which is a general perturbation based propagator that works directly with Two Line Elements (TLEs). We implement \texttt{sgpax} with the JAX \citep{jax2018github} differential programming package, giving us the ability to take arbitrary analytical derivatives between variables through auto-differentiation. Our code is based on the SGP4 implementation in \citep{vallado2001fundamentals,vallado2006revisiting} and follows the programming interface of the open source \texttt{python-sgp4} package \citep{sgp4py2012github}. \texttt{sgpax} is therefore a drop-in replacement for standard SGP4 propagators, allowing for effortless integration with other software pipelines.

Our SGP4 implementation is easily parallelised and can be run on Graphics Processing Units (GPUs) or tensor processing units (TPUs). This facilitates the use of large-scale Monte Carlo simulations, probabilistic inference, and applications in machine learning. Compared to \citep{acciarini2024closing}, the JAX backend provides a drop-in NumPy interface, better support for forward-mode automatic differentiation, and native auto-parallelisation support for Monte Carlo simulations, which are common in space mission planning. 

\subsection{Photometrically Accurate Orbital Simulation} \label{sec:render}

Given the dynamic range and intensity of illumination in LEO, photometrically accurate renderings are of key relevance when developing inspection missions~\citep{jawaid2025event, gentil2025mixing}. Taking advantage of the spectral rendering capabilities of Mitsuba 3~\citep{jakob2022mitsuba3}, we introduce a directional light source with spectral irradiance consistent with captured data~\citep{wehrli1985extraterrestrial}. The scene scale is in meters, providing radiometric intensity at the sensor in units of Watts per meter per steradian. While we focus this work on visual inspection, adjustments to the sensor spectral sensitivity would allow \dLITE{} to simulate inspections from sensors with different responses including infrared, multispectral, or hyperspectral cameras~\citep{felicatti2023hysim}.

\dLITE{} models dynamic lighting directions at each inspection point. The sun direction vector $\hat{\mathbf{l}}$ is defined as the negative unit vector towards the subsolar point, the point on the Earth with normal vector to the sun at the current orbital position and rotation of the Earth. This is computed using the solar azimuth~\citep{zhang2021solar}, based on the Equation of Time~\citep{hughes1989equation} to compute this point in Earth-Centred-Earth-Fixed coordinates. As these relations are defined by trigonometric polynomials and related to the Julian Day of the simulation, it is possible to pass gradients through the lighting vector.

A photorealistic model of the Earth was constructed with a custom Bi-directional Scattering Distribution Function (BSDF). Textures for the BSDF were taken from the NASA Blue Marble: The Next Generation~\citep{stockli2005blue_marble_next_generation} dataset and constructed to produce a similar Earthshine intensity, for accurate reflections of the background on the surface of the spacecraft. To match the visual complexity of the background, both the presence of clouds and atmospheric scattering are simulated at realistic altitudes.

\begin{figure}[htbp]
    \centering
    \includegraphics[width=0.8\linewidth]{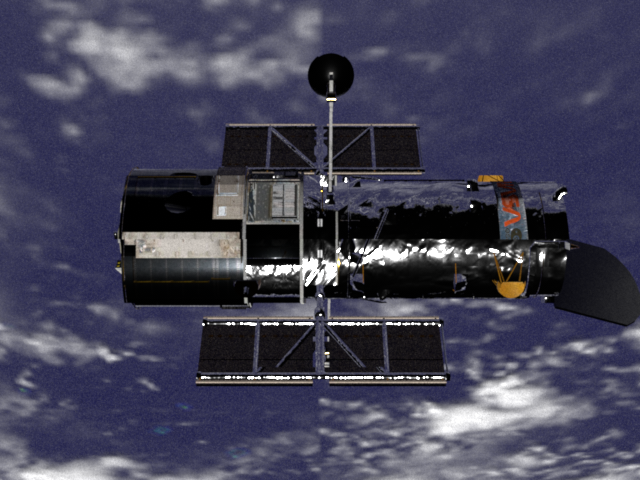}
    \caption{Simulated observation from our photometrically accurate simulation. Including photorealistic Earth models alongside representative visual models of a satellite, \dLITE{} produces high fidelity visual simulations of satellites in orbit.}
    \label{fig:visual_rendering}
\end{figure}

Images are rendered using a path replay backpropagation integrator suitable for handling volumetric scattering from the atmosphere~\citep{vicini2021path} and allowing derivatives to pass from cost maps in the image space back through poses of the camera and target satellite for optimisation. As the captured images are in radiometric units, we use a simplified sensor model to simulate them. This is required to be calibrated for a particular sensor, so for simplicity in our pipeline this is approximated using standard radiance-to-image plane transformations~\citep{Schott2007ImageChain} and indicative electron conversions to form digital numbers at the sensor. This component of our pipeline can be heavily customised to test custom image signal processing approaches and HDR imaging methods. An example image from our pipeline is shown in Figure~\ref{fig:visual_rendering}.

\subsection{Optimising for Imaging Conditions}\label{sec:vis_cost}
To demonstrate the capability of \dLITE{}, we show an example of how it may be used to improve imaging conditions in the context of bright specular reflections which are common in orbit. We emphasise that \dLITE{} is not limited to this application, as it gives the user the flexibility to specify sophisticated cost functions for diverse applications, so long as they can be expressed in terms of the outputs (e.g., images, physics parameters, etc.) with differentiable programming.

\textbf{Minimising Specularity}: Intense specular reflections pose one of the greatest challenges to the quality of data gained during an inspection. These regularly cause saturation of the sensor, blooming artifacts, and lens flares which can obscure key elements of interest such as textural and geometric changes. To benchmark \dLITE{}, we formulate a similar cost function to our prior work~\citep{mishra2024apaware} based on a Phong reflection model~\citep{phong1998illumination} to minimise specular reflections over an inspection period.

Given the modelled surface normals of the target satellite in the world frame $\hat{\mathbf{n}}$, and the sun illumination vector $\hat{\mathbf{l}}$, we compute the reflection vector $\hat{\mathbf{\omega}}_r$ as,
\begin{equation}
    \hat{\mathbf{\omega}}_r = 2(\hat{\mathbf{l}}\vdot\hat{\mathbf{n}})\hat{\mathbf{n}} - \hat{\mathbf{l}}.
\end{equation}
We consider an image $\mathcal{I} = \{\textbf{p}_{i,j}, M_{i,j}, \hat{\mathbf{n}}_{i,j} \}$ that is output from our renderer, where $\textbf{p}_{i,j}$ represents each pixel, $M_{i,j}$ is a binary mask for our specific target object, and $\hat{\mathbf{n}}_{i,j}$ is the normal vector for the surface that the pixel captures. We project out ray vectors for that pixel $\mathbf{\nu}(\mathbf{p}_{i,j})$, and define a cost function $\mathcal{L}_S$ for the specular intensity average over the pixels,
\begin{equation}
    \mathcal{L}_S (\mathcal{I})=  \frac{\sum_{{i,j}} M_{i,j}\mathrm{max}\qty(0, \hat{\mathbf{\omega}}_r \vdot \mathbf{\nu}(\mathbf{p}_{i,j}))^\alpha}{\sum_{{i,j}} M_{i,j}}, \label{eq:specular_cost}
\end{equation}
where $\alpha$ controls the width of the reflection lobe for the Phong reflection model and is material-dependent. We select an $\alpha=2$ which qualitatively matched the reflections from the render for our test platforms, but this value can either be based on measurements from the target or used as a hyperparameter to capture uncertainty in the reflection lobe.

\textbf{Maintaining Imaging Distance}: To ensure optimised relative orbits stay within a suitable imaging distance, we regularise the problem by introducing a distance-based cost. Assuming the initial orbit has a relative distance $d$ determined acceptable for imaging conditions, we introduce a simple $L_2$-loss for each pose,
\begin{equation}
    \mathcal{L}_d(r_c, r_t) = (||r_c-r_t||^2 - d)^2.
\end{equation}
We suggest that---depending on desired imaging quality requirements---similar cost functions could be evaluated directly from depth maps from the renderer while remaining differentiable, which would enable better performance for more complex geometries.

\section{Results and Discussion}
\subsection{Simulating Inspection Missions}

\begin{figure}[htbp]
    \centering
    \includegraphics[width=\linewidth]{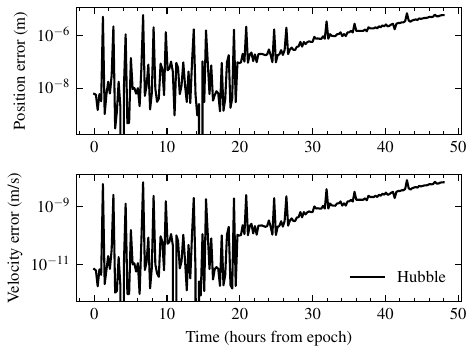}
    \caption{Error of our differentiable SGP4 propagator with respect to existing software~\citep{sgp4py2012github}, using the Hubble Space Telescope as an example. We achieve similar results, with the benefit of our implementation being fully differentiable via JAX~\citep{jax2018github}. Similar numerical results were observed for CloudSat and Sentinel-6 (not shown).}~\vspace{-\floatsep}
    \label{fig:test-error}
\end{figure}

We evaluated \dLITE{} by optimising inspection orbits for fictitious spacecraft monitoring each of the following three real-world satellites: Sentinel-6; CloudSat; and the Hubble Space Telescope.
We first validated our custom (differentiable) implementation of the SGP4 propagator by comparing it to existing (non-differentiable) implementations of the algorithm from \citep{sgp4py2012github}. Figure~\ref{fig:test-error} clearly demonstrates that our implementation closely matches that of \citep{sgp4py2012github} over 48\,hours of simulation time, with the slight increase in error over time attributed to the accumulation of floating-point inaccuracies.

\subsection{Trajectory Optimisation and Evaluation}
As mentioned in Section \ref{sec:opt_approach}, we optimise the chaser orbit for 200 iterations with an \texttt{Adam} optimiser. We run all experiments on a PC with an Intel i9-14900KF, an Nvidia RTX 4090 GPU, and 64GB of memory.

\begin{figure}[ht]
    \centering
    \includegraphics[width=\linewidth]{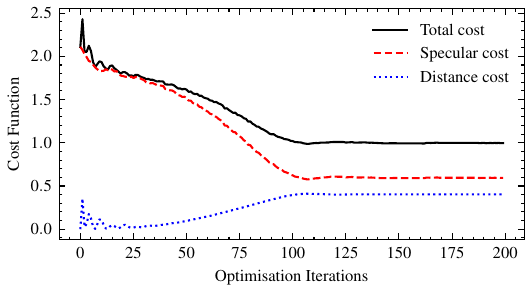}
    \caption{Total cost, specular cost ($\mathcal{L}_S$), and distance cost ($\mathcal{L}_d$) of the orbital parameters over optimisation iterations for the Hubble Inspection. Our optimisation reaches a local minimum yielding substantially reduced specular cost throughout an orbit.}
    \label{fig:cost_plot}
\end{figure}

Figure \ref{fig:cost_plot} shows the progression of the cost terms over the optimisation process. The curve shows rapid decrease in the total cost with convergence achieved at approximately $100$ iterations. We notice that the optimisation trades off deviating from the ideal imaging distance to lower specular costs. This trade-off can be tuned by re-weighting $\lambda_S$ and $\lambda_d$ according to the user's design specifications. 
Note also the rapid oscillations in the beginning of the optimisation process which eventually die down. These are indicative of poor conditioning of the cost landscape (e.g., due to deep and long but narrow valleys). Visual orbital inspection is particularly prone to this effect, as we may have diverse types of sophisticated cost functions in different units which our decision variables may affect at different scales. For example, a small change in eccentricity or inclination can put the chaser many kilometres away from the target. For this reason, we recommend either manual gradient preconditioning, scaling variables, or first order optimisers which involve some type of adaptive gradient normalisation such as \texttt{Adam} \citep{kingma2015adam}.

\begin{figure}[ht]
    \centering
    \includegraphics[width=\linewidth]{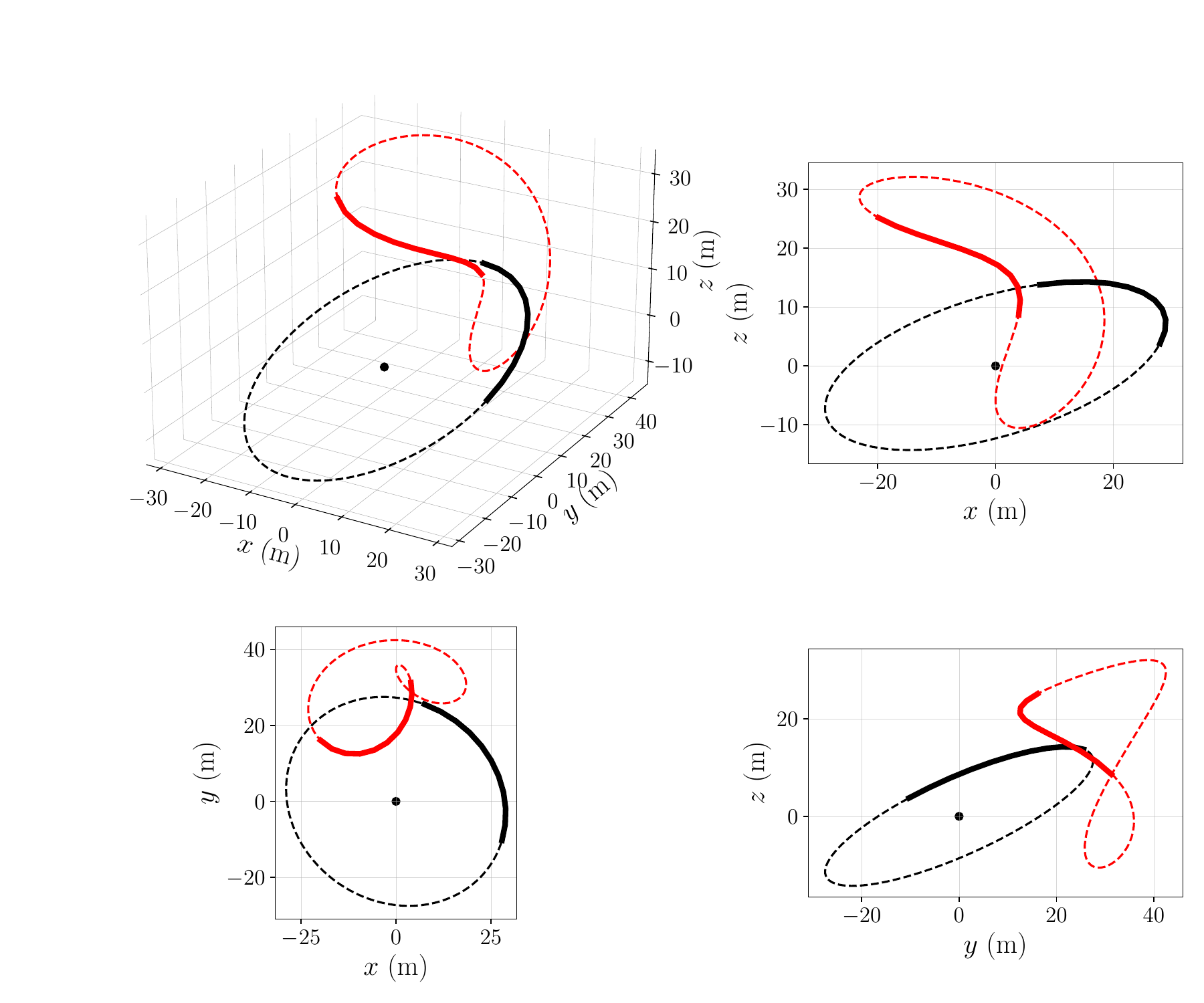}
    \caption{The orbit of the chaser satellite relative to the target satellite (black dot) in the Hubble inspection experiment. The black circle represents the initialised orbit and red curve represents the optimised orbit. The section of the orbit which the parameters were optimised for ($t\in[0,T])$ are solid, while the rest of the orbit is dashed.}
    \label{fig:relative_orbits}
\end{figure}

Figure \ref{fig:relative_orbits} shows the initial and the final orbits of the chaser in a frame relative to the target satellite, with the inspection period plotted in a solid curve. We notice that while the initial orbit is circular as designed, the final orbit is complex with an inherently 3D structure. It would be non-trivial to manually design such an inspection orbit to optimise the cost, which demonstrates the benefit of our methodology. We emphasise that the optimised (red) trajectory in Figure~\ref{fig:relative_orbits} is passively stable under the assumptions considered by SGP4. Note also that the inspection period (thick solid line) remains close to the desired imaging distance set in the cost function---other parts of the chaser orbit do deviate further from that distance, since they are not directly penalised in our cost function. If desired, this can be easily included by incorporating distance costs over the whole orbit rather than just the inspection period.

Table~\ref{tab:features_sat} demonstrates quantitative improvements in visual image quality using the proportion of saturated pixels in the satellite region~\citep{gentil2025mixing} and the inlier match ratio (MR) as metrics---i.e., inliers of the matched features over the total number of the keypoints detected~\citep{schonberger2017comparative} using SIFT features~\citep{lowe2004distinctive}. We select these metrics as proxies for image quality and 3D reconstruction, respectively. The proposed optimisation approach drastically reduces the proportion of the satellite body which is saturated in the image, whilst also boosting the number of SIFT features detected, compared to the conventional football inspection approach.

\begin{table}[!t]
\centering
\caption{Visual performance of the proposed optimisation through \dLITE{}. By optimising inspection trajectories, \dLITE{} significantly reduces the proportion of saturated pixels captured in each image and increases the proportion of inlier SIFT feature matches detected. Best values are in bold.}\label{tab:features_sat}
\begin{tabular}{l l r r}

Satellite & Approach & Sat. \% $\downarrow$& MR$\uparrow$\\
\hline
\multirow{2}{*}{HST} & Conventional & 26.0 & 0.759 \\
                            & \dLITE{} (Ours) & \textbf{20.5}  & \textbf{0.861}\\
                            \hline
\multirow{2}{*}{Sentinel-6} & Conventional & 1.20 & 0.583 \\
                            & \dLITE{} (Ours) & \textbf{0.04} & \textbf{0.632} \\
                            \hline
\multirow{2}{*}{CloudSat} & Conventional & 5.76 & 0.614 \\
                            & \dLITE{} (Ours) & \textbf{2.72} & \textbf{0.801} \\
\end{tabular}
\end{table}

\begin{figure}[htbp]
    \centering
    \includegraphics[width=0.95\linewidth]{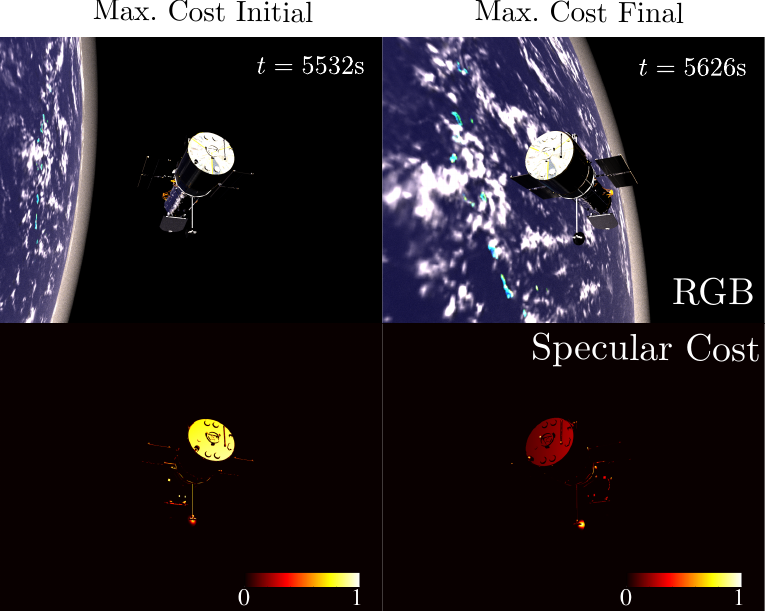}
    \caption{Qualitative comparison of maximum specular costs observed through an orbit. By optimising orbital parameters to observe the target satellite from different locations and angles, we dramatically reduce specular costs observed through an orbit.}
    \label{fig:qualitative_compare}
\end{figure}

In Figure~\ref{fig:qualitative_compare} we compare the qualitative results from \dLITE{} at the position of maximum cost. As can be seen, the overall specular cost in the final optimised orbit is significantly lower than that of the initial. Given the complex geometry of the satellite, it is not possible to avoid all specularities over all times, however by optimising across an inspection trajectory the overall impact can be significantly reduced.

\begin{figure}[t]
    \centering
    \includegraphics[width=\linewidth]{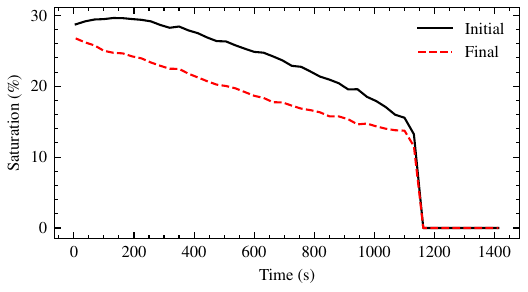}
    \caption{Percentage of saturated pixels on satellite body during an inspection of Hubble. \dLITE{} minimises our visual reflection cost function which is related to downstream metrics. It can substantially reduce the percentage of pixels which directly reflect sunlight, reducing saturation and thereby improving image quality.}
    \label{fig:image_saturation}
\end{figure}

To compare how cost minimisation translates to visual fidelity, Figure~\ref{fig:image_saturation} presents the percentage saturation of pixels observed on the satellite body throughout the orbit until eclipse. The final optimised orbit significantly reduces the proportion of saturated pixels across the inspection mission, validating the proposed specular cost as a viable means to improve image quality.

\subsection{Limitations}
Currently, \dLITE{} is limited to passive inspection trajectories with full knowledge of the target satellite state, and assumes that the camera on the chaser satellite always points directly at the target. We anticipate that combining variable attitude dynamics of both the target and chaser satellites in future work will enable non-obvious trajectories both in terms of orbital elements and sensor pointing. Moreover, while Mitsuba 3~\citep{jakob2022mitsuba3} allows for the customisation of simple pinhole cameras, these models neglect optical distortion, sensor point spread functions, and sensor noise which may be of interest in high-fidelity optical simulations~\citep{pajusalu2022sispo}. Motion blur is also neglected, meaning that \dLITE{} can only simulate accurate fly-by imagery~\citep{allworth2024use} for short exposure times or slow relative motion. Presently, we assume uniform reflection contribution from all materials on the satellite; however, different materials will have wider or narrower specular lobes. Calculating a Phong exponent $\alpha$ for each material with cost function in Equation~\ref{eq:specular_cost} would more accurately consider the specularities encountered on-orbit.

\section{Conclusions}

We have presented \dLITE{}---a fully differentiable pipeline for optimising on-orbit inspection trajectories from visual cost functions. We have demonstrated via simulation that by combining a differentiable rendering engine with a differentiable orbit propagator, we can construct relative inspection trajectories that avoid unfavourable visual features such as specular reflection, thereby improving the quality of captured images.

Future work with \dLITE{} will involve using the pipeline as a design tool for specific on-orbit inspection missions, where we will construct mission-specific cost functions to achieve favourable inspection trajectories. We will also include additional visual costs to minimise shadowing on the spacecraft body. By considering the quality of visual information in the optimisation of inspection trajectories, \dLITE{} achieves non-obvious relative orbits and substantially improved visual observations. 

\section*{Acknowledgements}
The authors gratefully acknowledge support from the NSW Space Research Network to present this work at the 76th International Astronautical Congress in Sydney, Australia. This research was supported in part through the NVIDIA Academic Grant Program. This work was supported in part by the ARC Research Hub in Intelligent Robotic Systems for Real-Time Asset Management (IH210100030).

\bibliographystyle{elsarticle-num} 
\bibliography{references}

\end{document}